\documentclass[conference]{IEEEtran}
\IEEEoverridecommandlockouts
\usepackage{cite}
\usepackage{amsmath,amssymb,amsfonts}
\usepackage{algorithmic}
\usepackage{graphicx}
\usepackage{textcomp}
\usepackage{xcolor}
\usepackage{multirow}
\usepackage{booktabs}
\usepackage{soul}
\usepackage{subcaption}
\usepackage{url}
\def\BibTeX{{\rm B\kern-.05em{\sc i\kern-.025em b}\kern-.08em
    T\kern-.1667em\lower.7ex\hbox{E}\kern-.125emX}}
\begin{document}

\title{Variational Voxel Pseudo Image Tracking}

\author{\IEEEauthorblockN{Illia Oleksiienko$^*$, Paraskevi Nousi$^\dag$, 
Nikolaos Passalis$^\dag$, Anastasios Tefas$^\dag$ and Alexandros Iosifidis$^*$}\\
\IEEEauthorblockA{\textit{$^*$DIGIT, Department of Electrical and Computer Engineering, Aarhus University, Denmark}\\
}
\IEEEauthorblockA{\textit{$^\dag$Department of Informatics, Aristotle University of Thessaloniki, Greece}}\\
\{io, ai\}@ece.au.dk \:\:\:\: \{paranous, passalis, tefas\}@csd.auth.gr
}

\maketitle

\begin{abstract}
Uncertainty estimation is an important task for critical problems, such as robotics and autonomous driving, because it allows creating statistically better perception models and signaling the model's certainty in its predictions to the decision method or a human supervisor.
In this paper, we propose a Variational Neural Network-based version of a Voxel Pseudo Image Tracking (VPIT) method for 3D Single Object Tracking. The Variational Feature Generation Network of the proposed Variational VPIT computes features for target and search regions and the corresponding uncertainties, which are later combined using an uncertainty-aware cross-correlation module in one of two ways: by computing similarity between the corresponding uncertainties and adding it to the regular cross-correlation values, or by penalizing the uncertain feature channels to increase influence of the certain features. In experiments, we show that both methods improve tracking performance, while penalization of uncertain features provides the best uncertainty quality.

\end{abstract}


\section{Introduction}

3D Singe Object Tracking (3D SOT) is the task of tracking an object in a 3D scene based on the given initial object position. This task combines challenges from both 3D Object Detection, as objects have to be accurately located in space, and 3D Multiple Object Tracking, as the object of interest has to be distinguished from similar objects. There is a variety of sensors that can be used for 3D SOT, including single or double camera setups, Lidar and Radar. While the camera setups are the cheapest option, they capture images which lack valuable for 3D SOT depth information, which can be provided by Lidar sensors. Lidars generate point clouds, which are sets of 3D points detected as the positions in the 3D scene of light beam reflections. The explicit depth information makes Lidar the most common choice for many 3D perception methods, including 3D SOT. The SOT is performed by predicting the offset of the object's position with respect to its previous known position. This has been approached by using correlation filters \cite{bolme2010tracking_filters, henriques2015kcf}, deep learning methods to directly predict the object's offset \cite{held2016goturn}, or by using Siamese methods which search for the position with the highest similarity score \cite{fang20213dsiamrpn,  bertinetto2016siamfc, li2018siamrpn, li2018siamrpnpp, oleksiienko2022vpit}. Since 3D perception methods are often used in critical fields, such as robotics or autonomous driving, it is important to provide accurate predictions and confidence estimations to avoid costly damages.

Uncertainty estimation in neural networks allows for using the network's outputs to better indicate the confidence in its predictions and to improve their statistical qualities, leading to better performance. The practical applications of uncertainty estimation are studied for several perception tasks, including 3D Object Detection \cite{feng20183ddropout, meyer2019lazernet, meyer2020ua3ddetlabels}, 3D Object Tracking \cite{zhong2020uavoxel, wang2020ua3dself}, 3D Human Pose Tracking \cite{daubney2011ua3dhuman}, and Steering Angle Prediction \cite{loquercio2020uncertaintydriving}. These methods provide an improvement in perception and control by using an uncertainty estimation process. However, most of these methods adopt single deterministic approaches to estimate different types of uncertainty, or use Monte Carlo Dropout (MCD) \cite{2016dropout} as an approach to estimate epistemic uncertainty. According to experiments in \cite{osband2021epistemic} on the uncertainty quality of different types of Bayesian Neural Networks (BNNs), MCD achieves the worst uncertainty quality.

In this paper, we introduce a Variational Neural Network (VNN) \cite{oleksiienko2022vnn} based version of the fastest 3D SOT method called Voxel Pseudo Image Tracking (VPIT) \cite{oleksiienko2022vpit} and propose two ways, i.e., the uncertainty similarity approach and the penalization approach, to utilize the estimated uncertainty and improve the tracking performance of the model.
The similarity-based approach computes a similarity between the estimated uncertainties to serve as an additional similarity score, while the penalization approach focuses on certain features by penalizing the feature values corresponding to high uncertainties. 
We train a VNN version of PointPillars for 3D Object Detection to serve as backbone for the proposed Variational VPIT (VVPIT) method. We, then, train the whole network following the VPIT's training procedure, but use the uncertainty-aware cross-correlation function and multiple samples of the Variational Feature Generation Network to compute uncertainty in the produced features. 
In experiments, we show that the use of uncertainty leads to an improvement in the model's tracking performance, and the choice of the penalty-based uncertainty utilization strategy leads to the highest improvement in Success and Precision metrics.

The remainder of the paper is structured as follows. Section \ref{S:RelatedWork} describes related and prior work. In Section \ref{S:ProposedMethod} we describe the proposed approach, including the Variational TANet model and its training and the proposed uncertainty-aware AB3DMOt. Section \ref{S:Experiments} outlines the experimental protocol and provides experimental results. Section \ref{S:Conclusions} concludes this paper\footnote{Our code is available at \url{gitlab.au.dk/maleci/opendr/vnn_vpit_opendr}}.

\section{Related Work}\label{S:RelatedWork}

Gawlikowski et al. \cite{gawlikowski2021uncertaintyindl} define four main categories of uncertainty estimation methods, based on the strategies they use to estimate the uncertainty of the model. Deterministic Methods \cite{2018evidentialdl, zhong2020uavoxel} use a single deterministic network and either predict its uncertainty by using an additional regression branch, or estimate it by analyzing the output of the model. Bayesian Neural Networks (BNNs) \cite{blundell2015weight, magris2023BNNsurvey} consider a distribution over weights of the network and compute the outputs of multiple model samples for the same input. The variance in the network's outputs expresses the estimated uncertainty, while the mean of outputs is used as the prediction value. Ensemble Methods \cite{osband2018randomized, valdenegro2019subens} consider a categorical distribution over the weights of the network and train multiple models at once. Test-Time Data Augmentation methods \cite{wang2018barintumortesttimeaug, wang2019aleamedical, kandel2021testtime} apply data augmentations commonly used in the training phase during the inference to pass distorted inputs to a single deterministic network and compute the variance in the model's outputs.

Variational Neural Networks \cite{oleksiienko2022vnn,oleksiienko2022vnntorchjax} are similar to BNNs, but instead of considering a distribution over weights, they place a Gaussian distribution over the outputs of each layer and estimate its mean and variance values by the corresponding sub-layers.
All types of uncertainty estimation methods, except those in the Deterministic Methods category, use multiple model passes to compute the variance in the network's outputs. 
This means the Deterministic Methods generally have the lowest computational impact on the model, but they usually perform worse than other methods. 
The single deterministic network approach can be improved by considering the Bayesian alternative, as it can be seen as a case of BNNs with the simple Dirac delta distribution over weights, which places the whole distributional mass on a single weight point.

The 3D SOT task is usually approached by using point-based Siamese networks, which consider a pair of target and search regions, predict a position of the target region inside the search region and compute the object offset relative to the previous object position.
P2B \cite{qi2020p2b}, BAT \cite{zheng2021box}, Point-Track-Transformer (PTT) \cite{shan2021ptt,jiayao2022real} and 3D-SiamRPN \cite{fang20213dsiamrpn} use point-wise Siamese networks and predict object positions based on the comparison of target and search point clouds. 3D Siam-2D \cite{zarzar2020efficient} uses one Siamese network in a 2D Birds-Eye-View (BEV) space to create fast object proposals and another Siamese network in 3D space to select the true object proposal and regress the bounding box. Voxel Pseudo Image Tracking (VPIT) \cite{oleksiienko2022vpit} uses voxel pseudo images in BEV space and deploys a SiamFC-like module \cite{bertinetto2016siamfc} to extract and compare features from target and search regions. Instead of using different scales, VPIT uses a multi-rotation search to find the correct vertical rotation angle.

Bayesian YOLO \cite{kraus2019bayesianyolo} is a 2D object detection method that estimates uncertainty by combining Monte Carlo Dropout (MCD) \cite{2016dropout} with a deterministic approach and predicts aleatoric uncertainty with a special regression branch, while computing the epistemic uncertainty from the variance in MCD model predictions.
Feng et al. \cite{feng20183ddropout} use a Lidar-based 3D object detection method and estimate the uncertainty in the predictions of the model in a similar way to Bayesian YOLO, by using a partially MCD model for the epistemic uncertainty estimation and using a separate regression branch for the aleatoric uncertainty estimation. LazerNet \cite{meyer2019lazernet} predicts the uncertainty of a 3D bounding box using a single deterministic network and utilizes the predicted uncertainty during the non-maximum suppression process. This approach is further improved by estimating the ground truth labels' uncertainty based on the IoU between the 3D bounding box and the convex hull of the enclosed point cloud, and using the provided uncertainties during the training process \cite{meyer2020ua3ddetlabels}. 

Zhong et al. \cite{zhong2020uavoxel} perform 3D Multiple Object Tracking (MOT) by using a single deterministic network for 3D Object Detection to predict the uncertainty in outputs and providing the estimated uncertainties to the tracker by replacing the unit-Gaussian measurement noise in Kalman filter \cite{1960kalmanfilter} with the predicted uncertainties.
Uncertainty-Aware Siamese Tracking (UAST) \cite{zhang2022uast} performs 2D single object tracking by using a single deterministic network and computing the distribution over the outputs by quantizing over the specific range of values and predicting the softmax score for each quantized value.
The final regression value is computed as an expectation of the corresponding quantized distribution, and the distributions are used to estimate better confidence scores and select the best box predictions.

To the best of our knowledge, there are no methods that utilize uncertainty for 3D Single Object Tracking. Moreover, the estimation of uncertainty for related tasks, such as 2D Single Object Tracking, 3D Multiple Object Tracking or 3D Object Detection, is based on single deterministic networks or MCD, despite the fact that the statistical quality of single deterministic networks can be improved by using a Bayesian alternative, and that MCD tends to produce the worst quality of uncertainty between BNNs \cite{osband2021epistemic}.

\section{Methodology}\label{S:ProposedMethod}

\begin{figure}[]
\centering
    \includegraphics[width=1\linewidth]{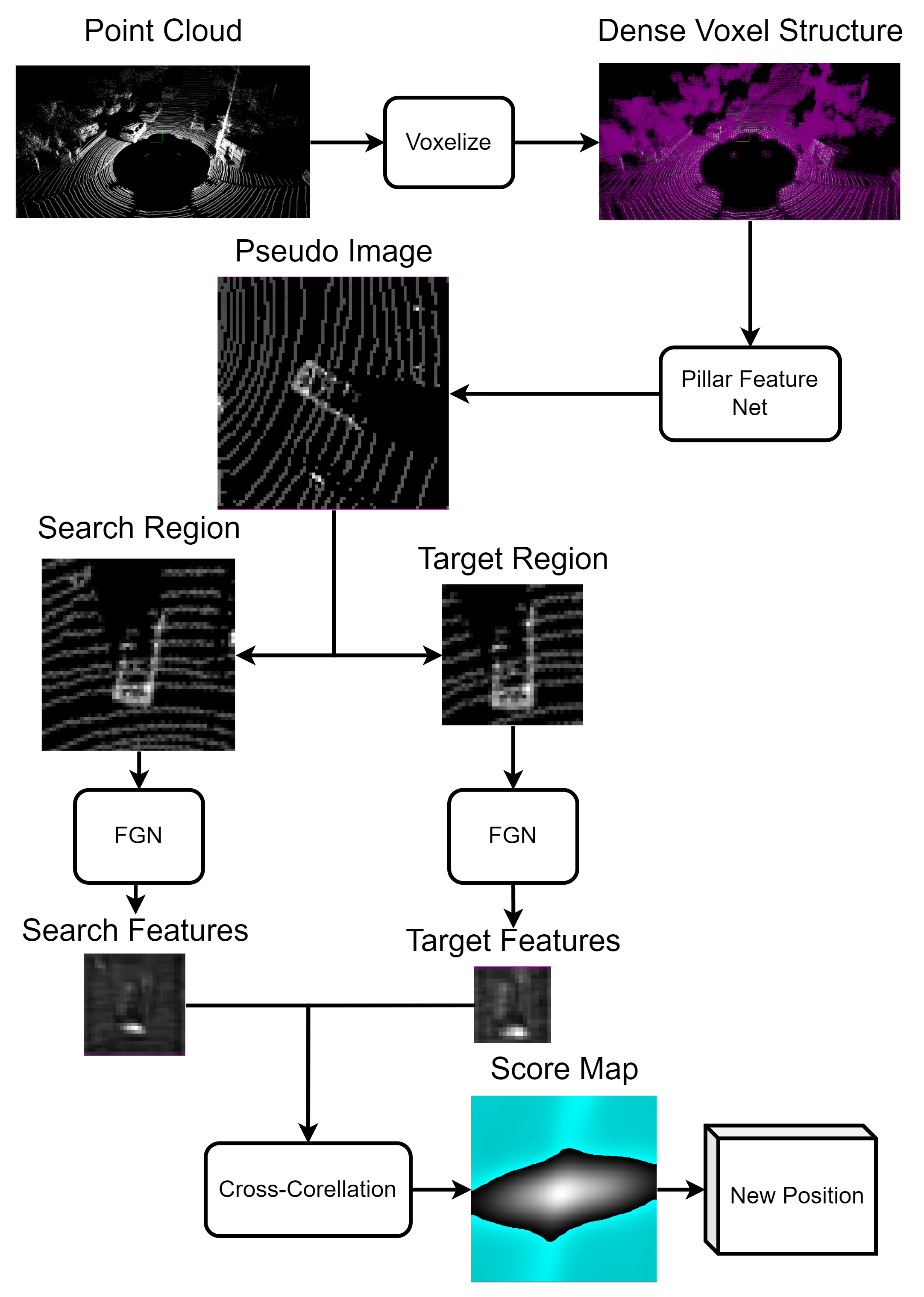}
    \caption{Voxel Pseudo Image Tracking structure.}
    \label{fig:vpit-structure}
\end{figure}

Voxel Pseudo Image Tracking (VPIT) uses PointPillars \cite{lang2018pointpillars} as a backbone to create voxel pseudo images and to process them with a Feature Generation Network (FGN), which consists of the convolutional part of the PointPillars' Region Proposal Network.
The search and target features are compared with a convolutional cross-correlation function that calculates a pixel-wise similarity map. The highest value in this similarity map is used to determine the object position offset between frames. The structure of VPIT is present on Fig. \ref{fig:vpit-structure}.

\begin{figure}
     \centering
     \begin{subfigure}[b]{0.3\linewidth}
         \centering
         \includegraphics[width=\linewidth]{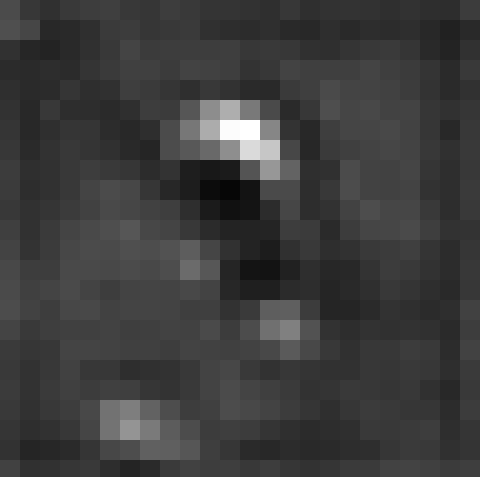}
         \caption{Mean}
     \end{subfigure}
     \hfill
     \begin{subfigure}[b]{0.3\linewidth}
         \centering
         \includegraphics[width=\linewidth]{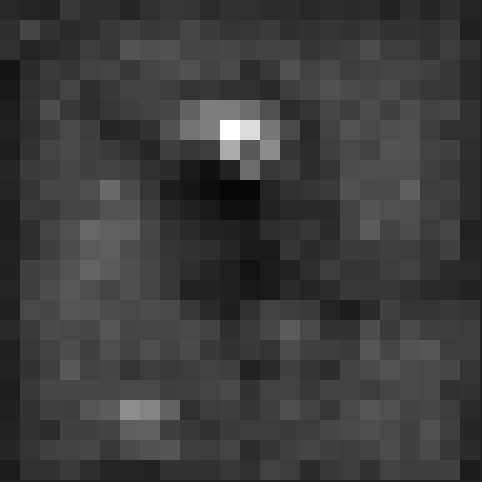}
         \caption{Variance}
     \end{subfigure}
     \hfill
     \begin{subfigure}[b]{0.3\linewidth}
         \centering
         \includegraphics[width=\linewidth]{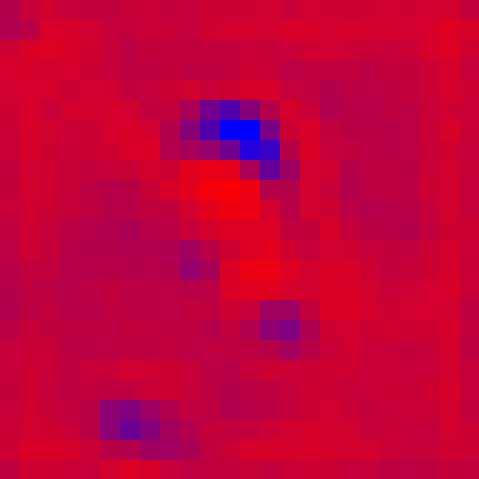}
         \caption{Mean \& certainty}
     \end{subfigure}
        \caption{An example of (a) mean, (b) variance and (c) mean and certainty features of a search region with a car in the center. Lighter color in the mean and variance images corresponds to higher values. Red color channel represents certainty in the corresponding pixel values, and blue color channel represents the mean feature values. The purple color indicates that the feature values and the certainty in those values is equally high, while the blue pixels signal features with high values and low certainty.}
        \label{fig:vvpit:mean-var-features}
\end{figure}

We train a Variational VPIT (VVPIT) by replacing the FGN subnetwork with a Variational Neural Network (VNN) \cite{oleksiienko2022vnn, oleksiienko2022vnntorchjax} based version of it, i.e., we create a Variational FGN (VFGN). We use multiple samples of the network for each input to compute mean and variance for the output features. The number of samples can be dynamic and is not required to be the same during training and inference. For each of target and search regions, VFGN produces a set of outputs in the form $Y = \{y_i, i \in [1, \dots, P]\}$ which correspond to the outputs of $P$ sampled VFGN models, with $Y^s = \{y^s_i, i \in [1, \dots, P]\}$ corresponding to the search region output set and $Y^t = \{y^t_i, i \in [1, \dots, P]\}$ to the target region output set.
The number of samples $P$ can be different for each set, but for simplicity, we use the same number of samples for both target and search regions.
The mean and variance of the outputs are computed as follows:
\begin{align}
\begin{split} \label{eq:scoremap}
    & y^s_m = \frac{1}{P} \sum^P_{i} y^s_i, \\
    & y^t_m = \frac{1}{P} \sum^P_{i} y^t_i, \\
    & y^s_v = \mathrm{diag}\left(\frac{1}{P} \sum^P_{i} (y^s_i - y^s_m)(y^s_i - y^s_m)^T\right), \\
    & y^t_v = \mathrm{diag}\left(\frac{1}{P} \sum^P_{i} (y^t_i - y^t_m)(y^t_i - y^t_m)^T\right), \\
\end{split}
\end{align}
where $y^s_m, y^s_v$ and $y^t_m, y^t_v$ are the mean and variance values of search and target output sets, respectively, and $\mathrm{diag}(\cdot)$ is a function that returns the main diagonal of a matrix.
Fig. \ref{fig:vvpit:mean-var-features} shows an example of the mean and variance values of features generated by the VFGN for a search region with a car in the center.
The background pixels have mostly high certainty, as all sampled models agree on them being irrelevant. The high magnitude features at the top part of the car have the highest uncertainty, as different model samples can disagree on the details in the appearance of the object.

The proposed VVPIT method can utilize the predicted uncertainties in different ways. The simplest way is to entirely ignore the uncertainty values and process the mean outputs only with the regular cross-correlation function $g(a, b)$, defined as a 2D convolution $\mathrm{conv2D}_{\omega=b}(a)$ with $\omega$ being the kernel weights.
This still leads to a statistically better model which can provide better predictions, but it can be further improved by utilizing the predicted uncertainties in the cross-correlation module.
Since most 3D SOT methods compare region features in a similarity manner, we focus on similarity-based approaches to use the uncertainty values, instead of applying distance-based approaches.
We propose a double similarity-based process to utilize uncertainty, which treats mean and variance values as separate feature sets and uses the convolutional similarity function $g(a, b)$ on both of them independently. The final similarity value $\hat{g}_{\mathrm{double}}$ is obtained by linearly the similarities of the mean and variance of the outputs as follows:
\begin{equation}
    \hat{g}_{\mathrm{double}}(y^s_m, y^t_m, y^s_v, y^t_v) = g(y^s_m, y^t_m) + \lambda g(y^s_v, y^t_v),
\end{equation}
where $\lambda$ is a variance weight hyperparameter. 
This approach is based on the idea that positions with similar uncertainties should be prioritized, as there is a high chance of them representing the same object.
Humans can also treat uncertainties as separate features.
Let us consider a task of classifying triangle and circle images, where some objects are rounded triangles.
Based on the deformation degree, people will have different values of aleatoric uncertainty in their predictions, as they will have harder time classifying rounded triangles as only one of the two classes.
If a person is asked to track these objects, the aleatoric uncertainty in predictions may be the only feature needed to distinguish between objects, given that size, thickness and other features are identical.
This is achieved by describing the tracked objects as ``definitely a circle", ``triangle with some curves", ``in between the circle and the triangle", which leads to low chances of mixing up these objects during tracking.
The same principle can be applied for Lidar-based 3D SOT task.
However, there are many different sources of uncertainty, considering the varying point cloud density, possible occlusions and object rotation.
Some parts of the object of interest may have uncertain features, and this uncertainty is likely to be preserved during the tracking process.

In addition to the above approach, we also define an uncertainty penalization process which places focus on features with higher certainty and penalizes the uncertain feature values.
This is achieved by dividing each mean feature value during the convolutional process by the corresponding normalized variance score, as follows: 
\begin{align}
\begin{split} \label{eq:scoremap}
    & \forall c, \:\: v^c_n(v) = (\rho - 1) \frac{v^c - \min(v^c)}{\max(v^c) - \min(v^c)} + 1, \\
    & \forall p_x, \forall p_y, \:\: \hat{g}_{\mathrm{pen}}(y^s_m, y^t_m, y^s_v, y^t_v)^{p_x, p_y} = \\
    &\hspace{3.5cm} = \frac{2 \Tilde{y^s_m}^{p_x, p_y} \Tilde{y^t_m}^{p_x, p_y}}{v_n(\Tilde{y^s_v})^{p_x, p_y} + v_n(\Tilde{y^t_v})^{p_x, p_y}},  
\end{split}
\end{align}
where the $v_n(v)$ function is used to normalize the variance predictions by the channel-wise minimum and maximum values to be in $[1, \rho]$ range, with a hyperparameter $\rho$ that defines how much the uncertain predictions are penalized, $v^c_n(v)$ implements the normalization procedure for a single channel $c$.
For an input $j$, $\Tilde{j}$ represents the tensor with convolutional patches of $j$, and $j^{p_x, p_y}$ corresponds to the values of $j$ at position $(p_x, p_y)$.

We follow the VPIT's training protocol and initialize a VVPIT model based on the VNN version of PointPillars for 3D Object Detection.
After the initialization, the model is trained with the Binary Cross-Entropy (BCE) loss between the ground truth and the predicted score maps.
Multiple VFGN samples are used during both training and inference to compute the mean and the variance in the target and search region features, which are later combined by using an uncertainty-aware cross-correlation module using one of the processes described above.

\section{Experiments}\label{S:Experiments}

We use the KITTI \cite{2012kitti} tracking dataset to train and test models.
Following the standard protocol, we use KITTI tracking training subset for both training and testing, as the test subset does not provide the initial ground truth positions.
The tracks $[0, \dots, 18]$ are used for training and validation, and tracks $19$ and $20$ are used to test the trained models.
Model performance is computed using the Precision and Success \cite{2016precision_success} metrics, which are based on the predicted and ground truth objects' center difference and 3D Intersection Over Union, respectively.
VPIT uses a pre-trained PointPillars network to initialize its pseudo image generation and FGN modules.
To follow the same procedure, we train a VNNs version of PointPillars on the KITTI \cite{2012kitti} detection dataset, use it to initialize the VPIT model and train the corresponding model for $64,000$ steps with different number of training VFGN samples per step in $[1, \dots, 20]$ range.

\begin{table}[]
\begin{center}
        \caption{Precision and Success values on the KITTI single object tracking experiments for VPIT and Variational VPIT (VVPIT) models.}
        \label{tab:vvpit:results-vnn}
        \centering
        \begin{tabular}{l|ccc}
            \toprule
            \textbf{Method} & \textbf{Uncertainty} & \textbf{Success} & \textbf{Precision}\\
            \midrule
            VPIT & - & 50.49 & 64.53 \\
            VVPIT & averaging & 51.97 & 66.69 \\
            VVPIT & double similarity & 52.62 & 66.56 \\
            VVPIT & uncertainty penalization & \textbf{53.30} & \textbf{67.79} \\
            \bottomrule
        \end{tabular}
\end{center}
\end{table}

Table \ref{tab:vvpit:results-vnn} contains the evaluation results of regular VPIT and the Variational VPIT (VVPIT) models with different ways to utilize the predicted uncertainty. 
We report the best-performing models for each uncertainty utilization process, which are obtained by using $20$ samples of the VFGN module.
By computing the average of predictions and discarding the variances, VVPIT achieves higher tracking performance compared to the VPIT model. By utilizing uncertainties, the Success and Precision values are further improved.
Both double similarity and uncertainty penalization processes lead to better models, but the penalization process leads to a better tracking performance.

\section{Conclusions}\label{S:Conclusions}
In this paper, we proposed a method to utilize uncertainty in 3D Single Object Tracking which uses a Variational Neural Network (VNN) based version of the VPIT 3D Single Object Tracking method to estimate uncertainty in target and search features and combines these features with an uncertainty-aware cross-correlation module.
We proposed two ways to utilize uncertainty in cross-correlation, i.e., by double similarity which adds a similarity in uncertainties to the regular cross-correlation, and by uncertainty penalization which penalizes uncertain features to shift focus to the more reliable feature channels.
Additionally, we tested the model's performance without exploiting the estimated uncertainties, as it still leads to a statistically better model compared to regular VPIT. 
The use of VNNs improves the tracking performance of VPIT in all cases, with the uncertainty penalization leading to the best Success and Precision values.

\section*{Acknowledgement}
This work has received funding from the European Union’s Horizon 2020 research and innovation programme under grant agreement No 871449 (OpenDR). This publication reflects the authors’ views only. The European Commission is not responsible for any use that may be made of the information it contains.

\bibliographystyle{IEEEbib}
\bibliography{bibliography.bib}

\end{document}